\documentclass[letterpaper, 10 pt, conference]{ieeeconf}  

\IEEEoverridecommandlockouts                              

\usepackage{booktabs}
\usepackage{graphics} 
\usepackage{epsfig} 
\usepackage{amsmath, amssymb}
\usepackage{balance}

\title{\LARGE \bf
A Study on Enhancing the Generalization Ability of Visuomotor Policies via Data Augmentation
}

\author{Hanwen Wang \quad 
\thanks{University of Chinese Academy of Sciences School of Artificial Intelligence \& Institute of Automation, Chinese Academy of Sciences \& NLPR}
}

\makeatletter
\def\@makecaption#1#2{%
  \vskip\abovecaptionskip
  \setbox\@tempboxa\hbox{{\normalfont\footnotesize #1: #2}}%
  \ifdim \wd\@tempboxa >\hsize
    {\normalfont\footnotesize #1: #2\par}
  \else
    \hbox to\hsize{\hfil\box\@tempboxa\hfil}
  \fi
  \vskip\belowcaptionskip}
\makeatother

\begin{document}

\maketitle
\thispagestyle{empty}
\pagestyle{empty}

\begin{abstract}

The generalization ability of visuomotor policy is crucial, as a good policy should be deployable across diverse scenarios. Some methods can collect large amounts of trajectory augmentation data to train more generalizable imitation learning policies, aimed at handling the random placement of objects on the scene's horizontal plane. However, the data generated by these methods still lack diversity, which limits the generalization ability of the trained policy. To address this, we investigate the performance of policies trained by existing methods across different scene layout factors via automate the data generation for those factors that significantly impact generalization. We have created a more extensively randomized dataset that can be efficiently and automatically generated with only a small amount of human demonstration. The dataset covers five types of manipulators and two types of grippers, incorporating extensive randomization factors such as camera pose, lighting conditions, tabletop texture, and table height across six manipulation tasks. We found that all of these factors influence the generalization ability of the policy. Applying any form of randomization enhances policy generalization, with diverse trajectories particularly effective in bridging visual gap. Notably, we investigated on low-cost manipulator the effect of the scene randomization proposed in this work on enhancing the generalization capability of visuomotor policies for zero-shot sim-to-real transfer.

\end{abstract}

\section{INTRODUCTION}
The generalization ability of visuomotor policies is crucial for robotic manipulation, as it determines whether the model can be effectively deployed. A good policy should be able to handle variations in object layouts, scene changes, and even enable cross-embodiment deployment. Existing approaches enhance the generalization of policies by collecting diverse datasets. Imitation learning methods based on human demonstrations have been proven effective in robotic unstructured manipulation tasks. Typically, human expert demonstration data is collected, which includes action signals generated by human teleoperation \cite{brohan2022rt, cheng2024open,fang2024airexo, fu2024mobile, iyer2024open, khazatsky2024droid, lin2024learning, shafiullah2023bringing, wu2024gello}. robot observation information such as RGB images from different viewpoints, and the robot's state (usually represented by the end-effector pose). Some studies have also collected task-related language instructions, enabling the policy to generalize object categories using pre-trained encoders from Vision-Language Models (VLM). Generally, imitation learning policies take visual images and the robot’s state as inputs and output the action the robot will perform for closed-loop position control \cite{driess2023palm, o2024open, black2024pi_0, chi2023diffusion, cheng2024open, kim2024openvla, kim2025fine, liu2025hybridvla, zheng2024tracevla, wang2025unified}. The cost in terms of both human effort and time for data collection is high; therefore, some methods attempt to leverage a small number of human demonstrations to generate large amounts of data for training more generalizable policies across scenes. These methods increase the diversity of trajectory data, enabling position generalization for objects on a desktop, and have achieved notable performance in tasks such as pick-and-place and simple assembly \cite{mandlekar2023mimicgen, deng2025graspvla, mu2025robotwin, nasiriany2024robocasa, jiang2024dexmimicgen}.

\begin{figure}
    \includegraphics[width=1\linewidth]{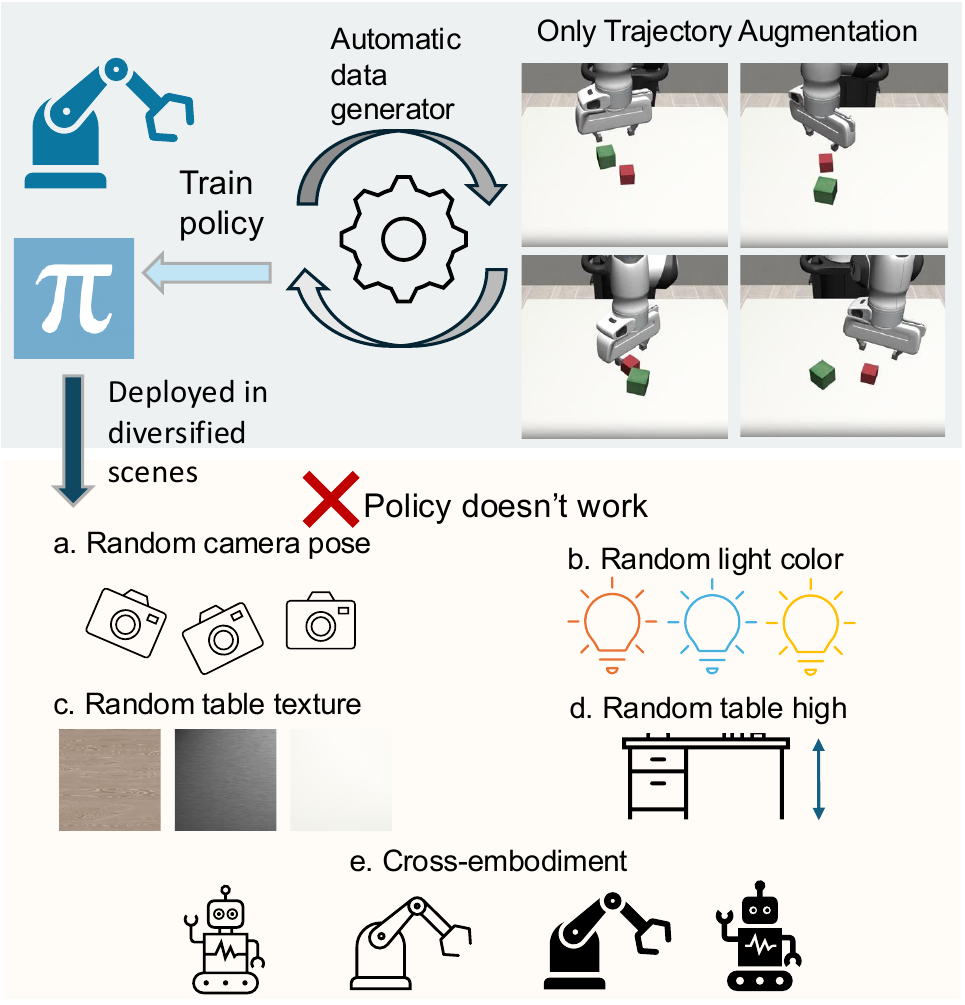}
    \caption{Previous work has primarily focused on trajectory augmentation methods. However, to obtain a  generalizable visuomotor policy, it is necessary to address other factors such as visual discrepancies. As the diversity of the scene increases, such as variations in camera poses, lighting colors, tabletop textures, table heights and cross-embodiment using data augmented by a single method becomes insufficient.}
    \label{fig1}
\end{figure}

However, prior work has primarily focused on automating the generation of diverse trajectories without accounting for visual discrepancies \cite{mandlekar2023mimicgen,nasiriany2024robocasa,jiang2024dexmimicgen}. For instance, methods like MimicGen \cite{mandlekar2023mimicgen} record the relative positions between the robot's end effector and the object to be manipulated based on a small number of human demonstrations. Even when the object's placement pose on a horizontal tabletop changes arbitrarily, the robot can still perform the task by continuously tracking both the object and the  end effector poses. Nonetheless, relying on this single-type data augmentation proves to be quite limited in enhancing the generalization capability of the policy. Some studies have explored the policy's performance under extensive domain randomization, aiming to bridge the visual gap and potentially facilitate sim-to-real policy deployment. Yet, these studies have not systematically assessed the impact of various domain randomization factors, and the conditions they investigate lack sufficient diversity. Moreover, cross-embodiment deployment using a single policy trained with such data has not been sufficiently explored.

Therefore, we systematically studied the impact of different domain randomization factors on the generalization ability of the policy. We focused on factors such as random camera poses, random lighting colors, random tabletop textures, and random table heights, and identified these as key factors influencing the policy's generalization capability. The data generated by existing methods does not cover these randomization factors. In addition, we explore how cross-embodiment data contribute to improving the generalization of visuomotor policies. For lighting and tabletop textures, we randomly sample between them. For camera poses, we ensured that each configuration sampled the same number of poses to maintain a balanced distribution. For cross-embodiment data, we used five manipulators with similar workspaces. To evaluate whether these scene randomization factors facilitate zero-shot sim-to-real deployment in real-world manipulation, we conducted experiments using a low-cost SO-101 manipulator. The results show that policies trained with scene randomization achieve stronger performance when objects are placed in more diverse configurations, effectively reducing the sim-to-real gap. Our contribution can be summarized as follows:
\begin{itemize}
\item[(1)] 
The limitations of different randomization factors on the generalization ability of the policy trained by previous methods were systematically studied.
\item[(2)] 
We systematically investigate whether different randomization augmentation factors can mutually enhance each other to improve the generalization performance of the policy.
\item[(3)] 
We conducted real-world experiments, confirming that policies trained with randomization augmentation facilitate zero-shot sim-to-real transfer.
\end{itemize}

\section{RELATED WORK}
\subsection{Manipulation data collection}
In the early stages, some works focused on collecting large-scale data for robotic learning. These efforts concentrated on self-supervised learning through trial and error to acquire robot grasping data\cite{levine2018learning, kalashnikov2018scalable, dasari2019robonet, pinto2016supersizing, kalashnikov2021mt}. Later, some works, such as GraspNet, did not collect robot trajectories but instead used analytical methods to label grasp poses directly \cite{fang2023anygrasp, wang20246}. The goal of robotic learning is not only to perform grasps but also to manipulate objects. Some related studies have collected large-scale operation datasets over extended periods of time using a large number of human operators \cite{mandlekar2018roboturk, jang2022bc, ahn2022can, ebert2021bridge}, but these methods still consume a significant amount of human and material resources. Some works on automatic data generation aim to address the problem of low-cost collection of large-scale trajectory data \cite{mandlekar2023mimicgen, jiang2024dexmimicgen, nasiriany2024robocasa, mu2025robotwin}. For example, MimicGen \cite{mandlekar2023mimicgen} efficiently generates a large number of trajectory data with only 10 human demonstration samples for a single task, which is essential for policy training. However, these automatic data generation methods focus solely on the diversity of trajectories, which still limits the generalization capability of imitation learning policies. The method we propose aims to expand the diversity of the data by adding variations such as different camera poses, lighting colors, and textures to address the lack of diversity in the dataset. Furthermore, while previous methods did not account for variations in the vertical placement of objects, we enhance the trajectory diversity by randomizing the tabletop height.

\subsection{Data augmentation in manipulation benchmarks}

Evaluating the generalization of a policy requires not only strong trajectory level generalization to handle diverse object configurations in the workspace, but also robust deployment under visual perturbations arising from variations in the scene. Some works have explored trajectory augmentation in simulators generating large amounts of trajectory data across various task settings \cite{jiang2024dexmimicgen}, \cite{jiang2024dexmimicgen}, \cite{mu2021maniskill}, \cite{james2020rlbench}, \cite{mees2022calvin}, but they largely overlook visual augmentation. Other works investigate combining trajectory augmentation with multiple visual randomization factors \cite{geng2025roboverse}, \cite{chen2025robotwin}. For example, RoboVerse \cite{geng2025roboverse} allows random variations in lighting conditions, background textures, and camera poses. However, these works have not systematically examined how such generalization factors affect the policy’s generalization ability, and their randomization settings remain limited. In many cases, camera poses are either manually predefined or only slightly perturbed, which restricts the generalization performance of policies trained with such augmented data. In contrast, our work introduces more principled randomization designs and systematically investigates the impact of different randomization factors on policy generalization.

\begin{figure*}
    \centering
    \includegraphics[width=\textwidth]{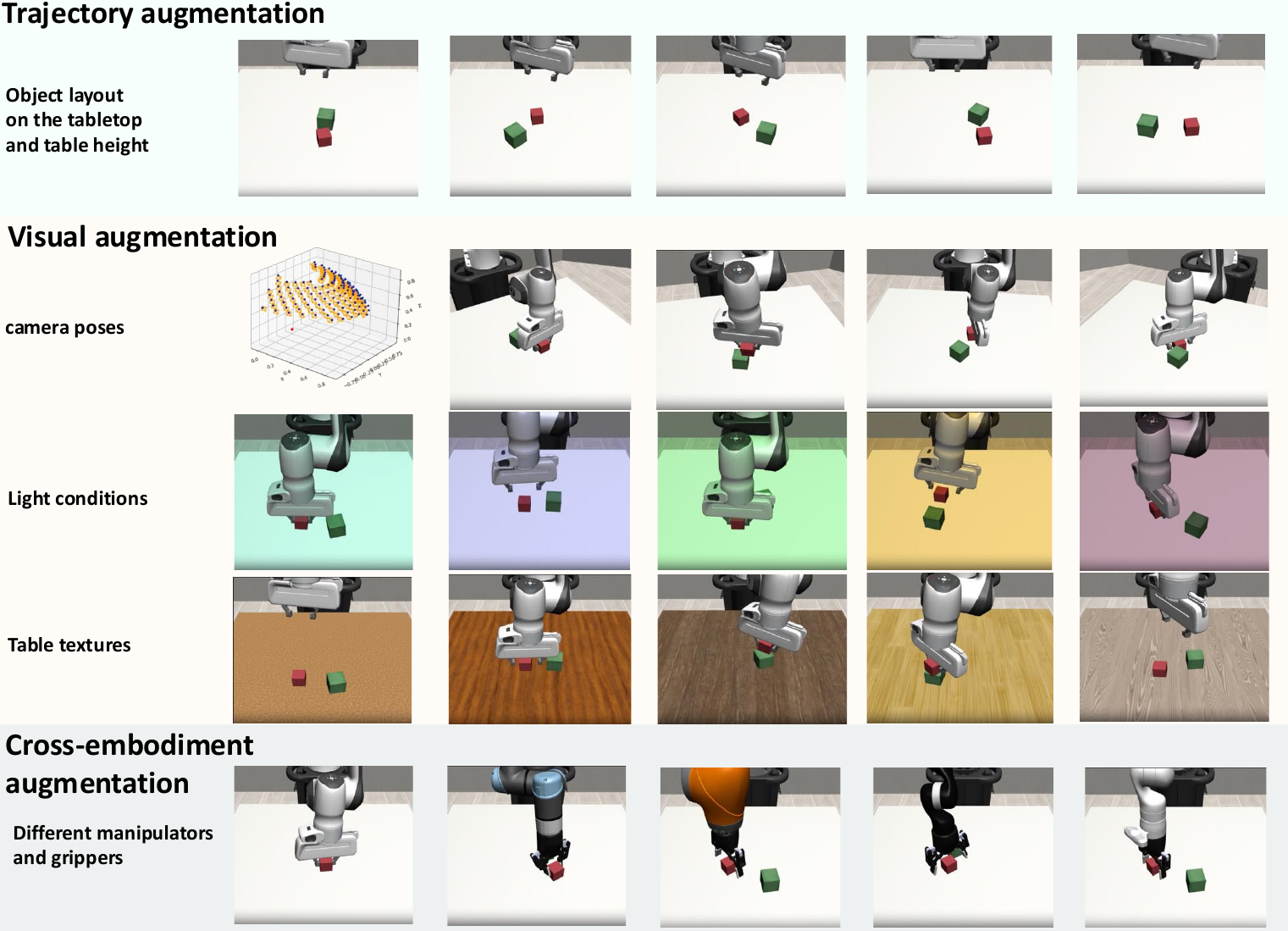}
    \caption{We focus on applying extensive scene randomization during the collection of robotic manipulation data in order to obtain a broad distribution of demonstrations. Compared to prior work, our trajectory augmentation introduces randomization over table height, as well as visual factors such as camera pose, lighting color, and tabletop texture. Furthermore, our data collection spans five types of manipulators and two types of grippers.}
    \label{fig2}
\end{figure*}

\section{PREREQUISITES}

\textbf{Policy learning.} All manipulation tasks in this paper are defined as a Markov Decision Process (MDP), where the robot needs to learn an observation based policy $\pi$ to model the relationship between visual inputs and robotic action outputs. The collected dataset consists of N demonstrations, denoted as $\mathcal{D}=\{(s_0^i,o_0^i,a_0^i,s_1^i,o_1^i,a_1^i,...,s_{H_i}^i)\}_{i=1}^N$, where $s\in\mathcal{S}$ denotes the state, $o\in\mathcal{O}$ denote the observations, $a\in\mathcal{A}$ denote the actions. At the beginning of each episode, the initial state $s_{0}^{i}$ is sampled from the distribution $D\subseteq S$. When employing imitation learning, we use Behavioral Cloning (BC) \cite{pomerleau1988alvinn}, which learns a policy by maximizing the likelihood objective.
\[
\arg\max_\theta\mathbb{E}_{(s,o,a)\sim\mathcal{D}}[\log\pi_\theta(a\mid s, o)].
\]
When training online reinforcement learning, we employ the generalized Proximal Policy Optimization (PPO) algorithm. Given an estimate of the advantage $\hat{A}(s,a)$, the PPO update is the sample approximation to

\begin{IEEEeqnarray*}{l}
\nabla_{\theta} \mathbb{E}_{(s_t,a_t)\sim \pi_{\theta_{\mathrm{old}}}} 
\min\Big(
\hat{A}^{\pi_{\theta_{\mathrm{old}}}}(s_t,a_t) 
\frac{\pi_\theta(a_t|s_t)}{\pi_{\theta_{\mathrm{old}}}(a_t|s_t)},\\
\hat{A}^{\pi_{\theta_{\mathrm{old}}}}(s_t,a_t) 
\operatorname{clip}\Big(\frac{\pi_\theta(a_t|s_t)}{\pi_{\theta_{\mathrm{old}}}(a_t|s_t)}, 1-\varepsilon, 1+\varepsilon\Big)
\Big)
\end{IEEEeqnarray*}

The clipping ratio $\varepsilon$ limits how much the policy can change from the previous policy.

\textbf{Assumptions.} The trajectory augmentation method we used during data collection is inspired by MimicGen \cite{mandlekar2023mimicgen}. Assumption 1: The action space A for each manipulator includes two components: a pose command to control the end effector and an actuation command for the gripper, which operates on a one-dimensional open/close basis. Assumption 2: We assume that a task can be decomposed into multiple object-centric subtasks, each defined with respect to the coordinate frame of the corresponding object, and that the sequence of subtasks is known. Assumption 3: During data collection, the pose of each object is assumed to be known at the beginning of every subtask.

\section{METHOD}
We focus on exploring the use of extensive data augmentations in the scene, As shown in Fig.\ref{fig2}, such that after policy learning, the policy acquires generalization capabilities. These generalization abilities manifest mainly as robustness to visual variations in the scene, resistance to background and lighting perturbations, the capability to manipulate objects arranged randomly on a tabletop, and the ability to be deployed across different embodiments. we describe how we introduce diverse randomization settings and present the imitation learning model we employ.

\subsection{Randomized Augmentation}

\textbf{Trajectory augmentation.} As stated in Assumption 2, each task can be decomposed into a sequence of object-centric subtasks, where each subtask is defined with respect to the coordinate frame of the manipulated object. Between subtasks, manually specified rule-based signals are available, which allow the automatic detection of subtask termination and thus enable the segmentation of each trajectory  into sub-trajectories $\{\tau_i\}_{i=1}^M$, with each $\tau_i$ corresponding to a distinct subtask $S_i$. Thus, each trajectory $\tau\in\mathcal{D}_{\text{src}}$ can be represented as a sequence of consecutive sub-trajectories, ${D}_{\text{src}}$ denotes the small set of demonstration data collected via human teleoperation. Specifically, During trajectory augmentation, suitable subtask segment are first selected from the pool of segments based on the pose of the manipulated object. Next, we treat the subtask segment $\tau_i$ as a sequence of end-effector poses in the world frame, $\tau_i=(T_W^{C_{i,0}},T_W^{C_{i,1}},\ldots,T_W^{C_{i,K}})$, where $C_t$ is the controller target pose frame at timestep $t$, $W$ is the world frame, and $K$ is the length of the segment. According to Assumption 3, even if the object is placed arbitrarily on the table, we can transform $\tau_i$ based on the object’s pose $T_W^{O_0}$ to obtain a new segment, $\tau_i^{\prime}=(T_W^{C_{i,0}^\prime},T_W^{C_{i,1}^\prime},\ldots,T_W^{C_{i,K}^\prime})$, where $T_W^{C_t^{\prime}}\:=\:T_W^{O_0}(T_W^{O_0})^{-1}T_W^{C_t}$. When generating a new trajectory segment, to ensure continuity, we adds an interpolation segment at the beginning of the transformed subtask segment. With this trajectory generation method, we can produce diverse trajectories simply by changing the placement of objects on the table. It is worth noting that previous work only randomized object positions on the tabletop without allowing vertical displacement, which greatly limits the generalization ability of the policy. To ensure that the policy can learn trajectories for objects at different heights, we vary the table height, allowing objects to be arranged at different heights relative to the robot’s coordinate frame.

\textbf{Visual augmentation.} We focus on three key factors: the pose of the third-person camera during data collection, lighting color, and tabletop texture, as shown in Fig.\ref{fig2}, When deploying a policy, deviations in camera pose from those used during data collection can lead to performance degradation, as such visual discrepancies reduce the policy’s generalization capability. To address this, we aim to expose the policy to a broader distribution of viewpoints during training.
Specifically, we construct a sphere centered at the robot base coordinate frame and then extract a portion of this sphere by removing viewpoints that are unlikely to be used during deployment and ensuring that the robot’s workspace is not heavily occluded. We then sample 100 camera poses uniformly on this surface using Fibonacci sampling. During data collection, to maintain a uniform distribution of viewpoints, we switch to the next camera pose only when trajectory augmentation results in a successful trajectory. For lighting condition randomization, we simulate dim indoor environments by randomizing the intensity of each RGB channel within the range of 0–0.5. For tabletop textures, we select 17 different patterns in RoboSuite \cite{robosuite2020} and the ARNOLD \cite{gong2023arnold}, including common wood and metallic textures.

\textbf{Cross-embodiment augmentation.}The robosuite simulator offers excellent scalability. Once a robot’s URDF file is available, trajectory and visual augmentation data collection can be easily deployed on robots with different configurations. As long as the robot’s workspace is not significantly different from that of the Franka Panda used in the original human demonstrations, deployment can generally be done directly, perhaps with minor adjustments to the placement of the robot base in space.
We selected commonly used laboratory robots such as the UR5e, IIWA, Kinova3, and Jaco. These robots have 6 or 7 degrees of freedom and are capable of performing manipulation tasks. For the end-effector, we chose the Robotiq85Gripper, which is also widely used in the robotics community. In the simulator, the Robotiq85Gripper is represented by six degrees of freedom. To better align its motions with those of the PandaGripper, we mapped the Robotiq85Gripper actions to a single opening/closing degree of freedom, which was also applied to the PandaGripper when feeding these actions as proprioceptive inputs during policy training.

\subsection{Imitation Learning Methodology}
We formulate policy learning within the framework of denoising diffusion probabilistic model (DDPM). A diffusion model represents a data distribution $p(x_0)$ via the reverse process of a forward noising process $q(x_k \mid x_{k-1})$, where Gaussian noise is iteratively added to $x_0$. The reverse denoising process is parameterized by a neural network $\epsilon_\theta(x_k, k)$ that predicts the injected noise. Sampling begins from $x_K \sim \mathcal{N}(0, I)$ and iteratively generates denoised samples following
\[
x_{k-1}\sim p_\theta(x_{k-1}\mid x_k):=\mathcal{N}\left(x_{k-1};\mu_k(x_k,\epsilon_\theta(x_k,k)),\sigma_k^2I\right),
\]
where $\mu_k(\cdot)$ is a deterministic function and $\sigma_k^2$ follows a fixed variance schedule. To leverage this framework for control, we adopt Diffusion Policy (DP) \cite{chi2023diffusion}, which conditions the denoising process on the current observation $s$. The policy parameterizes $p_\theta(a_{k-1}\mid a_k,s)$, and is trained via behavior cloning by fitting $\epsilon_\theta(a_k, s, k)$ to predict the noise injected into the action sequence. Unlike unimodal Gaussian policies, DP capture the multi-modal distribution of future trajectories without explicitly modeling $p_\theta(a_0 \mid s)$. For temporal consistency, the policy predicts an action chunk $a = {a_t, \dots, a_{t+T_p-1}}$ of horizon $T_p$, from which the robot executes $T_a \leq T_p$ steps before replanning. We adopt a CNN-based implementation of DP, which provides more stable training and requires minimal hyperparameter tuning.

\begin{figure}
    \includegraphics[width=1\linewidth]{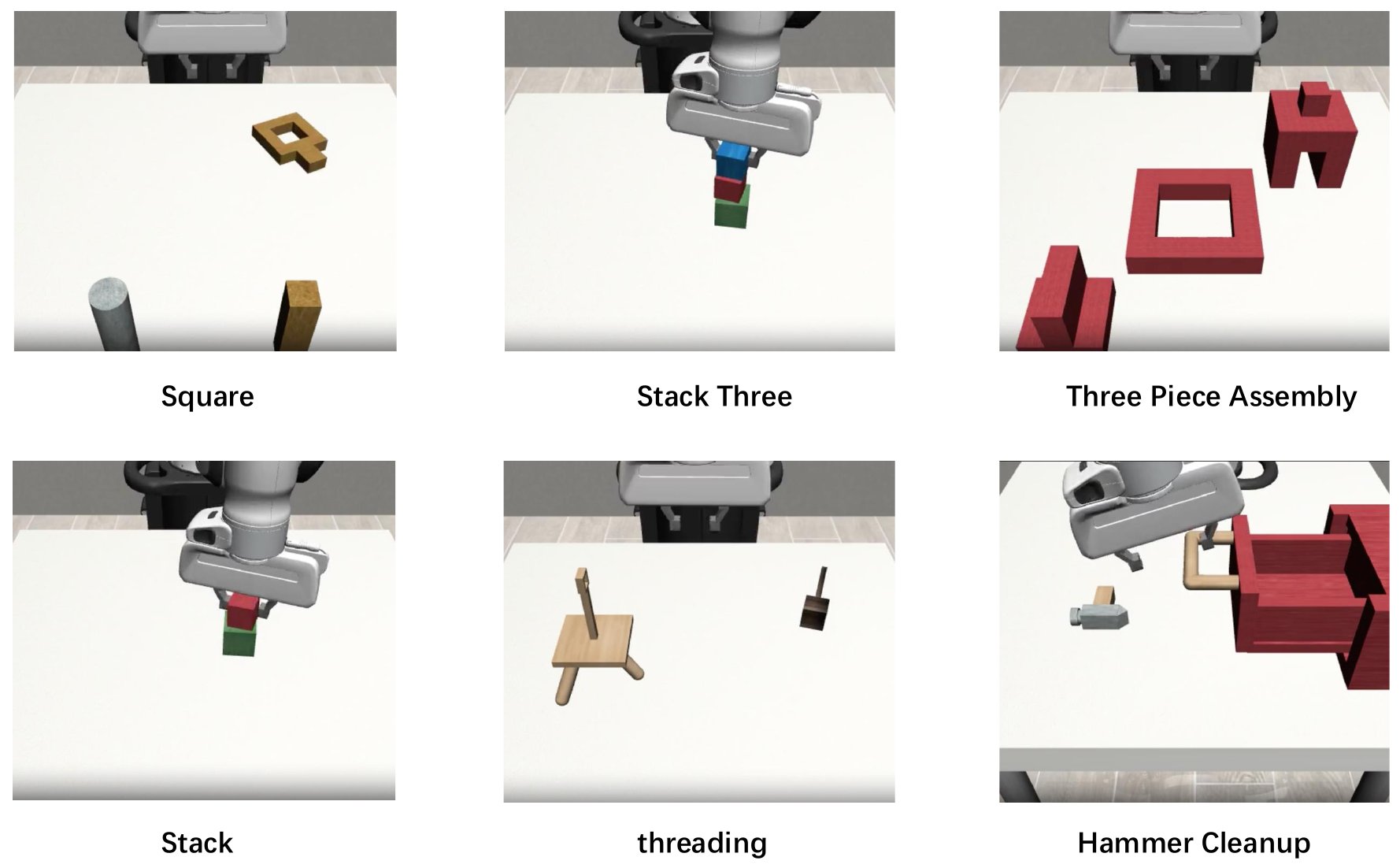}
    \caption{This work utilizes six robotic manipulation tasks from RoboMimic, focusing primarily on assembly or pick-and-place, which vary in difficulty.}
    \label{fig3}
\end{figure}

\begin{table*}[t]
\centering
\begin{tabular}{lllllll}
\toprule
Evaluation env  & Stack & StackThree & Square & HammerCleanup & Threading & ThreePieceAssembly \\
\midrule
Normal       & 1.00/-    & 0.76/-         & 0.58/-     & 0.96/-            & 0.64/-        & 0.68/-                 \\
Camera pose randomization     & 0.82/\textbf{1.00}      & 0.48/\textbf{0.78}  &  0.10/\textbf{0.60}   & 0.06/\textbf{1.00}   & 0.02/\textbf{0.72} & 0.00/\textbf{0.62}           \\
Light condition randomization & 1.00/0.94  & 0.50/\textbf{0.64}  & 0.46/\textbf{0.68}  & 0.92/\textbf{0.96}  & 0.26/\textbf{0.66} & 0.22/\textbf{0.68}          \\
Tabletop texture randomization   & 0.26/\textbf{0.90}      & 0.16/\textbf{0.74}  & 0.04/\textbf{0.48}   &  0.20/\textbf{1.00}  & 0.06/\textbf{0.58} & 0.02/\textbf{0.80}          \\
Table height randomization    & 0.96/0.94   & 0.58/\textbf{0.66}   &  0.42/\textbf{0.52}   &  0.58/\textbf{0.66}   &  0.10/\textbf{0.86}  & 0.36/\textbf{0.76}      \\
\bottomrule
\end{tabular}
\caption{Generalization of diffusion policies across randomization factors. Normal represents the original RoboMimic benchmark, The left side represents the performance of the model trained only with data generated by MimicGen as the baseline method, while the right side shows the performance of the model trained with data augmentation tailored to the corresponding test scenario.}
\label{tab1}
\end{table*}


\begin{table*}[]
\centering
\begin{tabular}{ccccccc}
\toprule
Evaluation env & w/o augmentation & CPA & LCA & TTA & THA  & CEA \\
\midrule
Camera pose randomization      & 0.25   & \textbf{0.77}   & 0.24    &  \textbf{0.28}  &  0.16  & \textbf{0.30}                 \\
Light condition randomization  & 0.56   & 0.28   & \textbf{0.76}    &  0.45  &  \textbf{0.63}  & 0.52                    \\
Tabletop texture randomization & 0.12   & 0.08   & \textbf{0.16}   &  \textbf{0.75}   &  \textbf{0.18}  & \textbf{0.20}               \\
Table height randomization     & 0.50   & 0.17   & \textbf{0.53}   &  0.39   & \textbf{ 0.73}  & \textbf{0.56}           \\

\bottomrule
\end{tabular}
\caption{The ablation study evaluates whether introducing additional randomization factors during training, under a single-factor setting, improves the policy’s generalization in test scenarios relative to a policy trained without augmentation. The table shows the average success rate over six tasks. CPA, LCA, TTA, THA, and CEA denote Camera Pose, Lighting Condition, Table Texture, Table Height, and Cross-Embodiment Augmentations, respectively.}
\label{tab2}
\end{table*}

\section{EXPERIMENT}
\subsection{Experiment setup}
We employed six tasks defined in the RoboMimic benchmark, as shown in Fig.\ref{fig3} and generated a large-scale dataset using trajectory augmentation, visual augmentation, and cross-agent augmentation. The dataset contains third-person and wrist-mounted RGB observations, proprioceptive states represented by the end-effector pose, and gripper states represented by the degree of opening. These synthetic data were used to train diffusion policies, which we conducted an in-depth investigation.

For real-world validation of the importance of randomization, we used a low-cost SO-101 manipulator, equipped with an Intel RealSense D435i camera. Online reinforcement learning in simulation was performed on an NVIDIA L20 GPU, while policy inference in real-world deployment was carried out on an NVIDIA GeForce RTX 3050 Ti GPU.

\subsection{Performance Analysis of Policies Trained with Augmentation}

 We designed a more challenging benchmark compared to RoboMimic. Specifically, instead of only altering the object layout in each scene, we also introduced a set of randomized factors of our own design. We first examined whether models trained with data collected under the original method could handle these more complex scenarios. Then, we trained new models with data augmented by our randomization strategies and investigated whether they could succeed under the same conditions.
As shown in Table \ref{tab1}, for each scenario we introduced only one randomized factor at a time, and compared the task success rates between the original method \cite{mandlekar2022matters} and ours. Each task was executed 50 rollouts with randomized scene configurations, and the success rate was reported. Experimental results indicate that models trained with the original method suffer from performance degradation once additional randomization factors are introduced. In contrast, models trained with our randomized augmentation demonstrate improved capability in these scenarios. These results validate the significance of studying the generalization ability of policies under such conditions, and further highlight that datasets generated by existing methods fail to sufficiently cover these challenging scenarios.


\subsection{Ablation Study}

We design a set of experiments to investigate how adding different randomization factors during data collection influences policy training, and whether these factors can promote complementary generalization performance across tasks. To investigate how different randomization factors contribute to the generalization of visuomotor policies, we design experiments where data is collected under scenes augmented with individual factors. Specifically, for each task, we introduce a single randomization factor to generate large-scale augmented trajectories. After training policies on these trajectories, we evaluate them in environments augmented with other randomization factors and report the task success rates. As shown in Table \ref{tab2}, The success rate reported in the table represents the average success rate across six tasks.

The experimental results indicate that these randomization factors has a mutually reinforcing effect on improving the generalization capability of the policy. after trajectory augmentation, such as training with height randomization and cross-embodiment data, the policy exhibits a certain degree of generalization even in visually randomized test environments. The cross-embodiment data inherently contain visual diversity, which further enhances this generalization capability. Although policies trained with visual augmentation show limited generalization in scenarios with randomized table heights, a policy trained under one type of visual randomization still demonstrates reasonable performance when evaluated under another visual randomization condition.

\begin{figure}
    \includegraphics[width=1\linewidth]{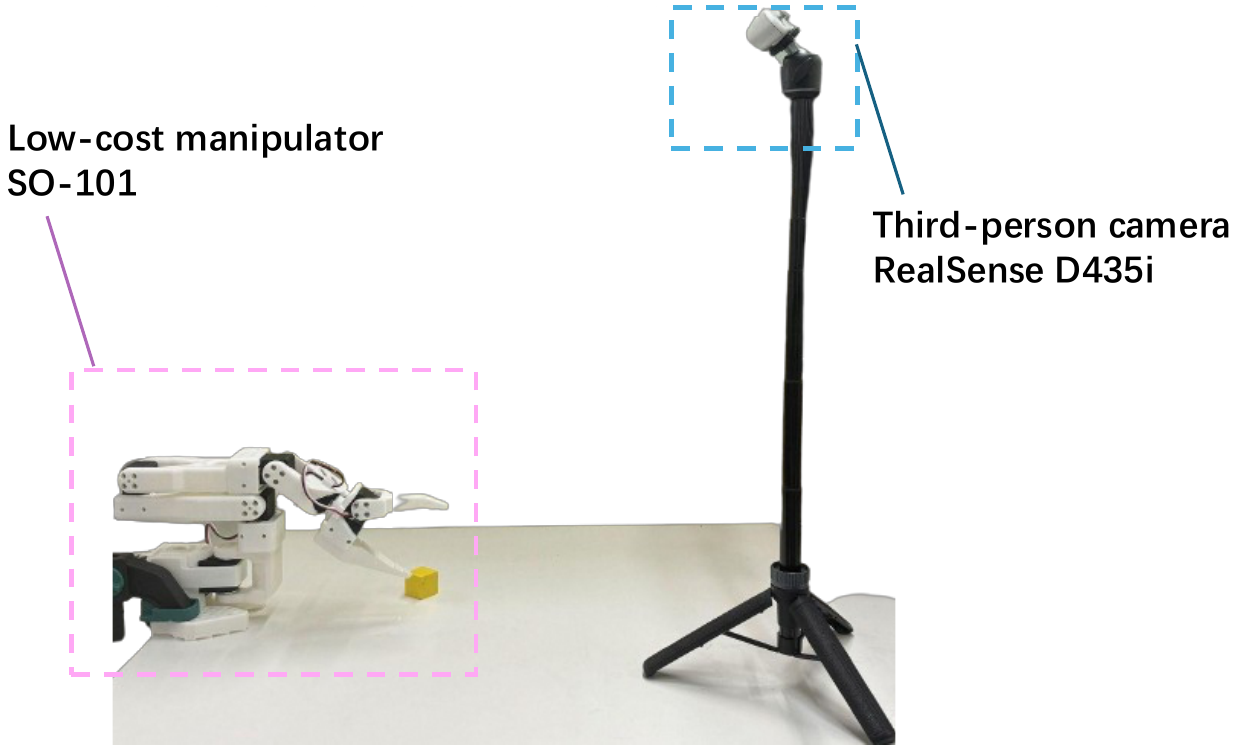}
    \caption{Our real-world experimental platform consists of a low-cost SO-101 manipulators on the left side and a RealSense camera placed on the right side from a third-person perspective.}
    \label{fig4}
\end{figure}

\subsection{Real-World Experiment}
To investigate whether introducing randomization during policy training facilitates zero-shot sim-to-real deployment, we set up a real-world experimental scene consisting of a low-cost SO-101 manipulator and a third-person Realsense D435i camera, as illustrated in Fig.\ref{fig4}. Following the Grasp Cube task defined in ManiSkill3 \cite{tao2024maniskill3}, we trained a multi-layer convolutional neural network policy with the PPO algorithm in simulation. In the training environment, we incorporated multiple randomization factors, including camera pose randomization, lighting condition randomization, object color randomization, and table height randomization. For comparison, we also trained a baseline policy in a separate environment without applying any randomization factors.

The experimental results, as shown in Table \ref{tab3}, demonstrate that policies trained with visual randomization achieve higher success rates and perform better when objects are placed over a wider range or with varying poses. Such enhancements in visual and trajectory diversity contribute significantly to enabling zero-shot sim-to-real robotic manipulation.

\section{Conclusion}
In this work, we systematically investigate the impact of scene and embodiment randomization on the generalization ability of visuomotor policies. By introducing trajectory augmentation and multiple visual randomization factors. On our proposed challenging benchmark, the policy trained with our method exhibits stronger generalization capability compared to prior approaches. We also found that the inclusion of these randomization factors has a mutually reinforcing effect on enhancing generalization capability. Real-world experiments show that randomized augmentation enhances the policy’s zero-shot sim-to-real deployment capability.

\section{Limitation}
If a policy is trained solely on the data provided in simulation in this work, it is difficult to perform complex manipulation tasks in the real world. In our simulation experiments, we did not explicitly align the simulated scenes with real-world scenarios, which results in a significant visual gap and some dynamics gap. Moreover, assembly tasks require not only diversity in visual and trajectory data but also an understanding of certain physical-world rules, which is challenging to achieve through purely vision-based imitation learning. A feasible approach is to first roughly align the visual configuration between the real and simulated environments, and reinforcement learning can be used to collect data while simultaneously applying both visual and trajectory augmentation., thereby acquiring more knowledge through interactions with the physical world.

\begin{table}[]
\centering
\begin{tabular}{ll}
\toprule
Training Env               & Grasp Cube \\
\midrule
w/o randomize augmentation & 0.28        \\
randomize augmentation     & \textbf{0.44}        \\
\bottomrule
\end{tabular}
\caption{Real-world experiment result.}
\label{tab3}
\end{table}





\balance
\bibliographystyle{IEEEtran}
\bibliography{Mybib}

\end{document}